%% file: ms.tex
\documentclass[10pt,twocolumn,a4paper]{article}

\usepackage[utf8]{inputenc}
\usepackage{url}
\usepackage{graphicx}
\usepackage{authblk}

\usepackage{algorithm}
\usepackage{algorithmic}

\usepackage{hyperref}

\usepackage{cite}
\usepackage{amsmath,amssymb,amsfonts, bm}
\usepackage{siunitx}
\usepackage{textcomp}
\usepackage{xcolor}
\usepackage[disable]{todonotes}
\usepackage[capitalise]{cleveref}
\crefformat{equation}{(#2#1#3)}
\crefrangeformat{equation}{(#3#1#4) to~(#5#2#6)}
\crefmultiformat{equation}{(#2#1#3)}{ and~(#2#1#3)}{, (#2#1#3)}{ and~(#2#1#3)}
\usepackage{pifont}
\include{abbreviations}
\setuptodonotes{inline}

\makeatletter
\def\blfootnote{\xdef\@thefnmark{}\@footnotetext}
\makeatother

\title{Surrogate-data-enriched Physics-Aware Neural Networks}
\author[1]{Raphael Leiteritz}
\author[2]{Patrick Buchfink}
\author[2]{Bernard Haasdonk}
\author[1]{Dirk Pfl\"uger}
\affil[1]{Institute for Parallel and Distributed Systems\\  University of Stuttgart, Germany}
\affil[2]{Institute of Applied Analysis and Numerical Simulation\\ University of Stuttgart, Germany}

\date{\vspace{-5ex}}

\begin{document}
\maketitle

\begin{abstract}
    Neural networks can be used as surrogates for PDE models.
    They can be made physics-aware by penalizing underlying equations or the conservation of physical properties in the loss function during training.
    Current approaches allow to additionally respect data from numerical simulations or experiments in the training process.
    However, this data is frequently expensive to obtain and thus only scarcely available for complex models.
    In this work, we investigate how physics-aware models can be enriched with computationally cheaper, but inexact, data from other surrogate models like Reduced-Order Models (ROMs).
    In order to avoid trusting too-low-fidelity surrogate solutions, we develop an approach that is sensitive to the error in inexact data.
    As a proof of concept, we consider the one-dimensional wave equation and show that the training accuracy is increased by two orders of magnitude when inexact data from ROMs is incorporated.
\end{abstract}



\section{Introduction}
Design, optimization or control of complex phenomena are tasks that are critical for applications such as CO$_2$ storage, e.g.\ \cite{Koppel2019}, or biomechanical simulations, e.g.\ \cite{biomech}. For computationally expensive high-fidelity simulations, such tasks are prohibitive. With computationally efficient surrogate models these tasks can be carried out in an approximative fashion.


Machine Learning techniques can be used to derive such surrogates. Neural networks are one class of such data-driven methods that have successfully been applied to the solution of partial differential equations (PDEs) in various settings \cite{nn_pde1, nn_pde2,pinn}.
In a recent work \cite{neural_op}, it has even been demonstrated that a data-driven method can outperform a numerical method in both, accuracy and runtime, to solve an inverse uncertainty quantification problem.

Already a few decades ago, methods have been proposed that use neural networks to solve PDEs by constraining the loss function with the underlying equations \cite{nn_pde1}.
In recent years, this original idea has seen a renaissance in the form of so-called Physics-Informed Neural Networks (PINNs) \cite{pinn}.
These have meanwhile been successfully applied to a variety of problems such as reconstructing pressure and velocities from visual flow data, simulating blood-flow in cardiovascular structures \cite{pinn_hfm} or subsurface flow \cite{pinn_subsurface}.
In contrast to methods learning directly on simulation data \cite{nn_pde_data} using e.g.\ CNNs \cite{nn_pde_cnn} or LSTMs \cite{dl_lstm}, PINNs add a term to the loss function which penalizes predictions that do not satisfy the underlying PDE.
In the scope of this paper, we differentiate these loss terms by their nature.
The \emph{physical (or BVP) loss} aims to minimize (a) the PDE residual on interior data points and (b) the (initial) boundary data on boundary data points.
PINNs featuring an additional \emph{data loss} on interior data points are referred to as \emph{data-enriched} PINNs.

So far data-enriched PINNs in literature are based on expensive measurements which are either obtained from real or numerical experiments.
An example is \cite{pinn} where PINNs are trained on experimental data and afterwards are used to estimate model parameters of the PDE to solve this inverse problem.

\begin{figure}[htbp]
    \centerline{\includegraphics[scale=.5]{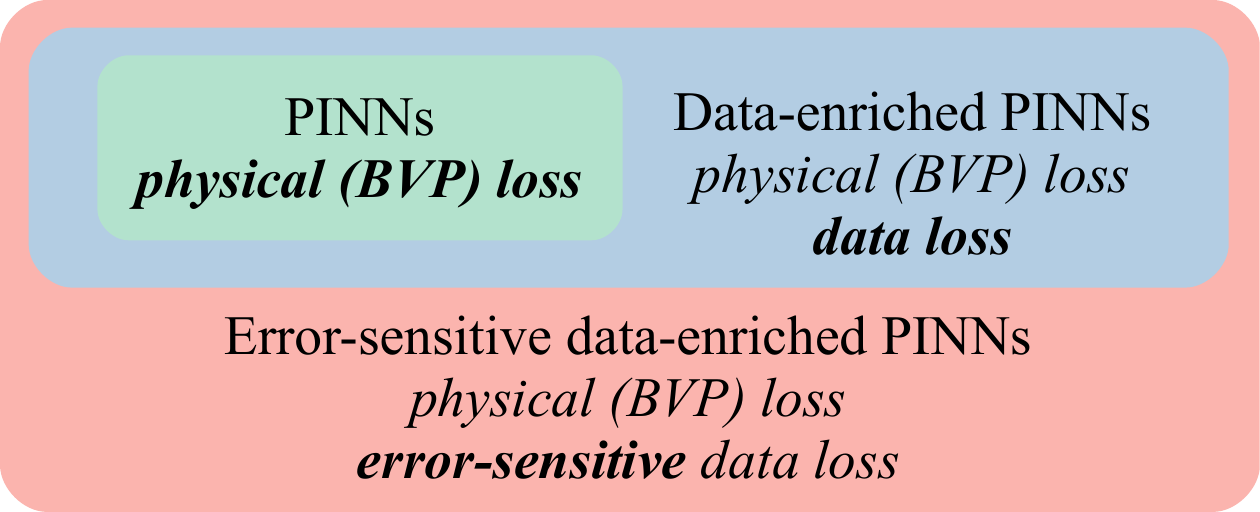}}
    \caption{Illustration of different PINN approaches within this paper.}
    \label{fig:workflow}
\end{figure}

In this work, we aim to integrate comparably cheap data from surrogate models in the data loss of the PINN. In a slightly more general view, we call this data \emph{inexact data} as solutions from surrogates are approximative solutions. We additionally assume that this inexact data is provided with an error bound that quantifies the error with respect to the exact solution. This setting is quite natural for many surrogate models like e.g.\ Reduced-Order Models (ROMs).
As our main contribution, we propose the notion of \emph{error-sensitive} PINNs (see \cref{fig:workflow}) as a generalization of data-enriched PINNs.
The core idea is that the error bound from the inexact data is taken into account during the training by relaxing the optimization goal if the error with respect to the inexact data is smaller than the error bound.
This approach comes with two key advantages:
Firstly, the additional knowledge on the solution within the solution domain may provide a boost to training times as well as prediction accuracy as it now offers the optimizer more data to find a correct solution.
This is crucial since, in a simulation setting, data is usually scarce due to their high computational costs.
Secondly, since surrogates can be of low fidelity, the error-sensitive part does not force the PINN to fit the inexact data but instead gives it leeway to improve over the inexact data.
Thereby, PINNs are encouraged to refine the given inexact data.

Previous studies for PINNs have shown \cite{lra2020,vdM2021} that the prediction accuracy is sensitive to the weighting of the different loss terms.
In order to apply \cite{vdM2021} to data-enriched PINNs, we generalize this idea to the case of more than two loss terms. Moreover, we include a comparison of the different weightings in our numerical experiments.

The performance of PINNs on inexact data and the new error-sensitive approach are studied in a numerical experiment based on the one-dimensional linear wave equation. We show that the error-sensitive PINN outperforms a standard non-data-enriched PINN. Moreover, the experiments show that weighting the losses correctly is essential.

The rest of the paper is structured as follows: In \cref{sec:pre}, we introduce the essentials of scientific machine learning for PDEs using PINNs. Subsequently, we present the loss weighting strategies, the error-sensitive PINNs and ROM-data-enriched PINNs in \cref{sec:data_based_loss}. The new methods are compared to classical approaches in \cref{sec:experiments} in a numerical experiment for the one-dimensional linear wave equation. \cref{seq:conclusion} concludes the paper and provides an outlook to future work.

\section{Prerequisites}
\label{sec:pre}

For the neural network architecture, we restrict ourselves to conventional fully-connected neural networks.
\todo{subscript i of f is not defined?}
We introduce these as a concatenation of $\nlayers$ different layers $f_i$
\begin{equation}
    \fnet(x;\theta) := (\flayer_{\nlayers} \circ \ldots \circ \flayer_2 \circ \flayer_1)(x)
\end{equation}
where $\theta \in \R^{\nTheta}$ represents the vector of all trainable parameters, such as the weights and biases.

The neural networks are tailored towards a specific goal by minimizing a cost functional, the so-called loss (functional), $\approxLoss(\theta): \R^{\nTheta} \rightarrow \R_{\geq 0}$ over the set of all possible network parameters~$\theta$.
Typically, some form of stochastic gradient descent method such as ADAM \cite{adam} is used for the optimization of $\approxLoss(\theta)$.
The classical example for a loss to learn an input--output mapping from $\nData$ input--output pairs $\{(\xData, \yData)\}_{i=1}^{\nData}$ is
\begin{align}\label{eq:loss_data}
    \approxLoss(\theta) = \ldata(\theta) := \frac{1}{\nData} \sum_{i=1}^{\nData} \left( \yData - \fnet(\xData; \theta) \right)^2 ,
\end{align}
which is known as Mean Squared Error (MSE) loss. In the following, we call this loss term the \emph{data loss}.


The Physics-Informed Neural Networks (PINNs) modify the loss function to inform the network about the underlying physics \cite{pinn}. In the scope of this paper, the desired physical property is a boundary value problem (BVP) of the following type: find an unknown function $u: \domain \rightarrow \R$ with $\domain \subset \R^d$ such that
\begin{align}\label{eq:bvp}
    \diffOp[u] = & \; 0 \quad \text{in } \domain,          &
    \bdryOp[u] = & \; 0 \quad \text{on } \partial \domain,
\end{align}
where $\diffOp$ is some, potentially non-linear, differential operator and $\bdryOp$ is an operator prescribing the boundary conditions.
In the scope of this paper, time-dependent problems are of particular interest.
In that case, $x=(t,\xi) \in \domain := \domainT \times \domainXi$ is composed of the time $t$ and a spatial coordinate $\xi$.
Compared to numerical simulations, which aim to ensure that the laws of physics are not violated, a PINN does not strictly guarantee a physically valid solution.
Instead, it encourages the solution $\fnet(\theta) \approx u$ to satisfy the BVP in selected collocation points $\xInterior$, $\xBoundary$ with
\begin{align}\label{eq:loss_interior_boundary}
    \begin{split}
        \linterior(\theta)
        :=&\; \frac{1}{\ninterior} \sum_{i=1}^{\ninterior} \lb \diffOp[\fnet(\theta)](\xInterior) \rb^2\\
        \lboundary(\theta)
        :=&\; \frac{1}{\nboundary} \sum_{i=1}^{\nboundary} \lb \bdryOp[\fnet(\theta)](\xBoundary) \rb^2\\
    \end{split}
\end{align}
and adds these terms as penalty terms to the loss function where the derivatives in $\diffOp$ are evaluated via Automatic Differentiation~\cite{Baydin2018}. The PINN loss then reads
\begin{align}
                                                            & \approxLoss(\theta)
    = \sum_{j \in \idxSet} \lambda_j \approxLoss_j(\theta), &
                                                            & \idxSet := \{ \dataLabel, \interiorLabel, \boundaryLabel \},
\end{align}
with weights $\lambda_j \in \R_{\geq 0}$ and the loss contributions $\approxLoss_j(\theta) \geq 0$ from  \cref{eq:loss_data,eq:loss_interior_boundary}. We call $\linterior(\theta)$ the interior or PDE residual loss and $\lboundary(\theta)$ the boundary loss. Both these terms together are referred to as the BVP losses.

\section{Data-enriched PINNs}
\label{sec:data_based_loss}
Theoretically, PINNs work without the data loss ($\lambdaData = 0$). For \emph{data-enriched} PINNs ($\lambdaData > 0$), we would like to additionally use $\yData = u(\xData)$, but the exact solution $u$ is frequently not available or too expensive to compute. In the scope of this paper, we investigate how data-enriched PINNs behave, if the target function in the data loss is provided by inexact data, e.g.\ with an approximate solution $\yData = \uROM(\xData) \approx u(\xData)$.

\subsection{Loss Weighting for PINNs}
\label{sec:weighting}
For the case of non-data-enriched PINNs, it has been observed in previous studies that the choice of weights $\lambda_j$ in the loss function is crucial for the training speed and quality \cite{lra2020,vdM2021}.


The \emph{Learning Rate Annealing for PINNs} (LRA) in \cite{lra2020} is motivated by a stiffness-phenomenon in the gradient flow dynamics.
It uses the statistic of the gradient to balance the interplay of all loss contributions $\approxLoss_j$.
The \emph{Optimal Loss Weight} (OPT) in \cite{vdM2021} is a heuristic approach that tries to balance the losses based on the assumption that the relative error in the derivatives of the neural network can be bounded uniformly. We generalize this idea to more than two losses. In this formulation it chooses the loss based on characteristic quantities $M_j$ of the loss $\approxLoss_j$ for each $j \in \idxSet$, e.g.\ $\Mdata[u] \approx \LtDomainNorm{u}^2 / \mabs{\domain}$ for the data loss. The resulting weights are
\begin{align*}
    \lambda_j[u] = \lb \sum_{k \in \idxSet} M_j[u] / M_k[u] \rb^{-1}.
\end{align*}
Note that the factors $M_j[u]$ may depend on the exact solution which is not available. In our numerical experiment, we compute the weights from the exact solution for the sake of simplicity. For practical applications however, one could use the inexact data $\uROM$ to compute the factors.


\subsection{Error-Sensitive PINNs}
\begin{figure}[htbp]
    \centerline{\includegraphics[scale=.5]{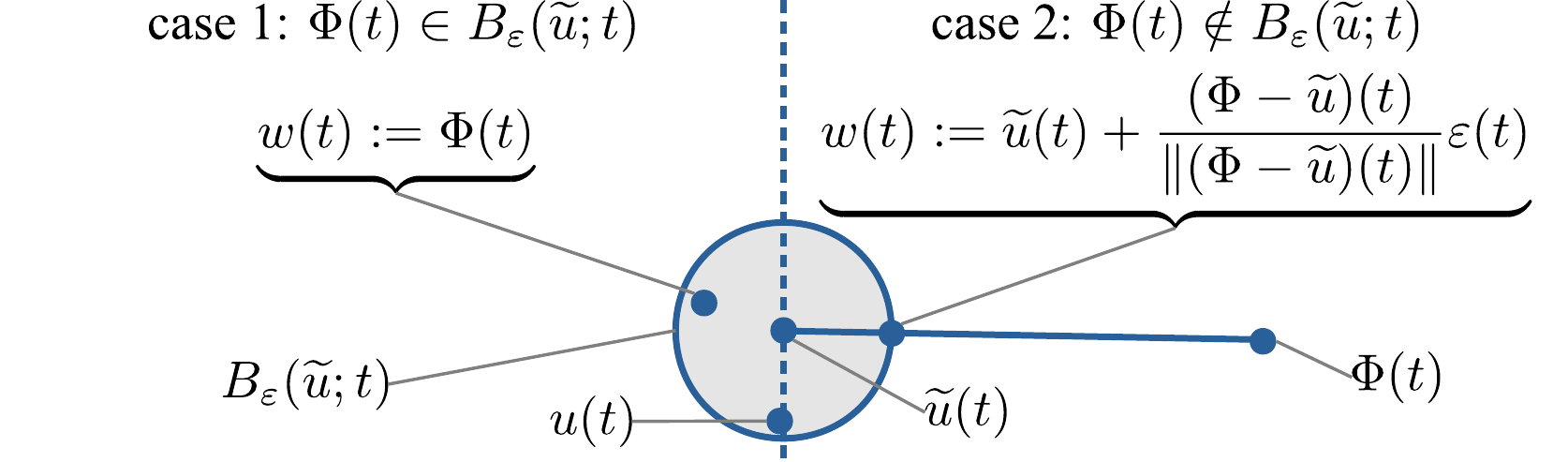}}
    \caption{Illustration of the definition of $w(t)$.}
    \label{fig:ball}
\end{figure}

The following section focuses on time-dependent problems with $x = (t, \xi)$ in the sense of \cref{sec:pre}. For all functions $w(x)$, we define the short-hand notation $w(t) := w(t, \cdot)$ for each $t \in \domainT$.
The analysis is formulated in terms of a non-discrete analogue to the data loss from \cref{eq:loss_data},
\begin{align*}
    \Ldata(\theta)
    = \LtDomainNorm{u - \fnet(\theta)}^2,
\end{align*}
where Monte-Carlo integration is used to approximate $\Ldata(\theta) / \mabs{\domain} \approx \ldata(\theta)$ with $\nData$ sampling points $\xData \in \domain$ and $\yData = u(\xData)$ for $1 \leq i \leq \nData$. Moreover, the data loss $\Ldata(\theta)$ is sampled separately in time and space. For the sake of simplicity, we consider an equidistant sampling in space and time in the following, e.g.\ for $\xi \in \domainXi \subset \R$ and $\dt,\,\dxi > 0$,
\begin{align*}
    t_i =   & \; t_0 + i\dt,    &
    \xi_j = & \; \xi_0 + j\dxi, &
\end{align*}
with $1 \leq j \leq \nXi$, $1 \leq i \leq \nT$.
If we would know the exact solution $u$, the equidistantly sampled data loss would read
\begin{equation}
    \label{eq:ldata}
    \begin{aligned}
        \ldata(\theta)
        := & \; \frac{1}{\nT} \sum_{i=1}^{\nT} \lb \ixi[u](t_i; \theta) \rb^2,                    \\
        \lb \ixi[u](t_i; \theta) \rb^2
        := & \; \frac{1}{\nXi} \sum_{j=1}^{\nXi} \mabs{(\fnet(\cdot; \theta) - u)(t_i, \xi_j)}^2.
    \end{aligned}
\end{equation}
As $u$ is not available, we use the inexact data $\uROM$ instead, for which we assume that the error can be quantified for each $t \in \domainT$ with
\begin{align}\label{eq:surrogateErr}
    \LtNorm{u(t) - \uROM(t)}{\domainXi} \leq \errBound(t)
\end{align}
with an error bound $\errBound: \domainT \rightarrow \R_+$.
The idea of the \emph{error-sensitive} data-enrichment is to trust the inexact data only up to the error bound $\errBound(t)$. To this end, consider the open $\errBound(t)$-ball around $\uROM(t)$
\begin{align*}
    \ball := \big\{ w \in L_2(\domainXi) \big| \LtDomainXiNorm{w - \uROM(t)} < \varepsilon(t) \big\}
\end{align*}
which can also be interpreted as a tube around $\uROM$ over time $t$.
In order to differentiate whether $\fnet(t)$ lies in $\ball$, we define $w: \domainT \rightarrow \closedBall$, the projection of $\fnet(t)$ onto the closed $\errBound(t)$-ball around $\uROM(t)$, see \cref{fig:ball}.

The data loss $\Ldata(\theta)$ can then be bounded with
\begin{align*}
    \Ldata(\theta)
    \leq & \; 2 \lb \LtDomainNorm{u - w(t)}^2 + \LtDomainNorm{w(t) - \fnet}^2 \rb.
\end{align*}
The first term on the right side, can be bounded with $4\errBound(t)^2 \mabs{\domainT}$ since both, $u(t)$ and $w(t)$, are elements in the ball $\ball$ from which follows that their distance is bounded by the diameter of the ball, $2\errBound$. This yields
\begin{equation}
    \label{eq:es_estimate}
    \begin{aligned}
        \Ldata(\theta) / \mabs{\domain}
        \leq                  & \; 2 \LdataWrom(\theta)  / \mabs{\domain} + 8\errBound(t)^2 / \mabs{\domainXi} \\
        \LdataWrom(\theta) := & \;
        \int_{\domainT}
        \underbrace{
            \LtDomainXiNorm{w(t) - \fnet(t; \theta)}^2
        }_{
            =: \lb \IxidataWrom(t; \theta) \rb^2
        } \,\intd t
    \end{aligned}
\end{equation}
which describes how the error from \cref{eq:surrogateErr} propagates through the training. Moreover, this estimate guarantees that training with the error-sensitive loss $\LdataWrom(\theta)$ improves the networks quality with respect to the original data loss $\Ldata(\theta)$ (as long as $\errBound(t)$ is small enough).

Due to the choice of $w(t)$, the term $\IxidataWrom(t; \theta)$ in \cref{eq:es_estimate} is (a) zero if the error-sensitive PINN solution $\fnet(t; \theta)$ is in $\ball$ and (b) equal to the distance between $\fnet(t; \theta)$ and the closest point on the boundary $\partial\ball$ otherwise. This can be expressed with
\begin{align*}
    \IxidataWrom(t; \theta)
    =  & \; \ReLu\lb\LtDomainXiNorm{\fnet(t; \theta) - \uROM(t)} - \errBound(t)\rb, \\
    \ReLu(x)
    := & \; \begin{cases}
        0, & x < 0    \\
        x, & x \geq 0
    \end{cases}.
\end{align*}
This results in the the error-sensitive loss
\begin{align*}
    \LdataHeur(\theta) / \mabs{\domain}
    \approx & \; \ldataHeur(\theta)
    := \frac{1}{\nT} \sum_{i=1}^{\nT} \lb \ixidataWrom(t_i; \theta) \rb^2,                    \\
    \ixidataWrom(t; \theta)
    :=      & \; \ReLu\lb \ixi[\uROM](t; \theta) - \errBound(t) / \mabs{\domainXi}^{1/2} \rb.
\end{align*}

Note that the presented loss is readily implementable in the major ML frameworks due to usage of the well-established $\ReLu$ function. Moreover, the non-error-sensitive approach is a special case of error-sensitive data-enrichment if the data is fully trusted, i.e.\ $\errBound \equiv 0$.

\subsection{ROM-data-enriched PINNs}
\label{subsec:MOR}
A prominent example for inexact data certified with an error bound is data obtained from so-called Reduced Order Models (ROMs). ROMs are constructed to flexibly trade accuracy for efficiency by choosing different sizes of the reduced basis. At the same time many ROMs provide an error bound that can be evaluated efficiently. An example for ROMs with a time-dependent error bound of the assumed form \cref{eq:surrogateErr} is derived in \cite{Glas2020} for the linear wave equation. Technically, we additionally assume that the underlying FEM approximation space is rich enough to approximate $u$ and thus, that the error between the FEM solution and the exact solution is negligible.

\section{Experiments}
\label{sec:experiments}
\begin{figure}[htbp]
    \centerline{\includegraphics[width=0.9\columnwidth]{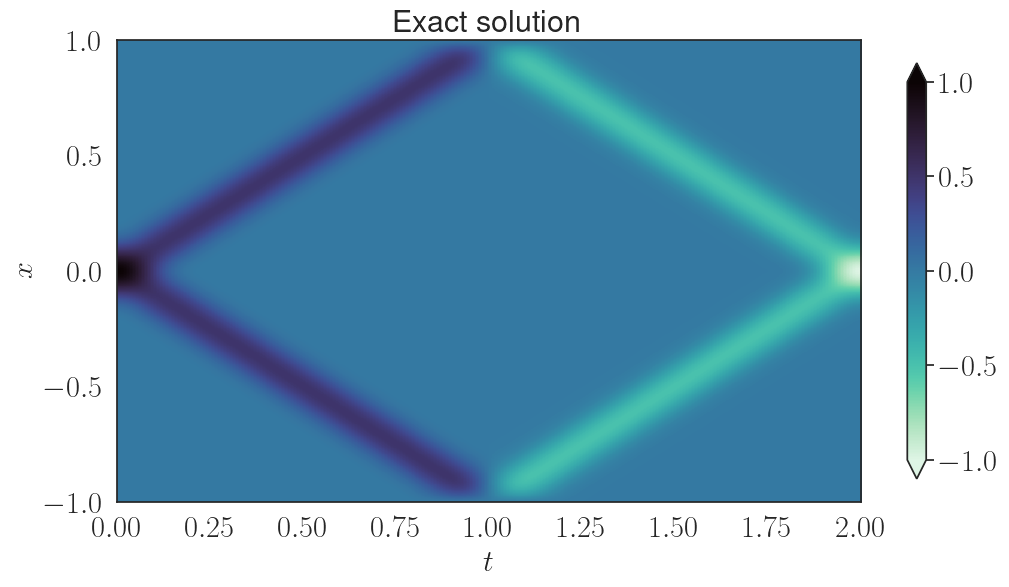}}
    \caption{Visualization of the exact solution. A single bump in the middle of the domain at $t=0$ travels outwards and is reflected at the boundaries at $t=1$.}
    \label{fig:solution}
\end{figure}

We consider the wave equation in a one-dimensional spatial domain
$\domainXi := (-1, 1)$ over the time interval $\domainT:=(0, 2)$, for
which an analytical solution is available for validation. The PDE on
the spatio-temporal domain $\domain := \domainT \times \domainXi$
is
\begin{equation*}
    \partial^2_t u(t, \xi) - \partial^2_\xi u(t, \xi) =\; 0 \qquad
    \left(t, \xi\right) \in \domain
\end{equation*}
with homogeneous Dirichlet boundary conditions and zero initial velocity $v_0\equiv 0$. The initial data and the solution are visualized in \cref{fig:solution}.


\todo{better: relative error}
The different approaches are compared with respect to the mean squared error
\begin{equation}
    \MSE{y} := \LtDomainNorm{u - y}^2 / \mabs{\domain}.
\end{equation}
Additionally, experiments are repeated $r$ times and averaged to account for random initial weight configurations which we denote with $\mMSE{r}{\fnet}$.
\todo{this notation of displaying different weight initializations in very non-standard (?). One very bad run might ruin everything. What about median or mean of log-quantities? For mean of log-quantities, consider the relation with the geometric mean}



The network used is a conventional fully connected feed-forward network with \emph{tanh} activation functions. Its architecture and hyperparameters were chosen with a hyperparameter optimization for the non-data-enriched PINN over $407$ individual runs resulting in $\nlayers = 5$ layers, $\nneurons = 20$ neurons per layer, and a learning rate of $\alpha = \num{1e-3}$.
The optimizer is ADAM with default parameters. The network parameters are initialized using the truncated Xavier initialization \cite{init}.
The number of sampling points of the different losses varies for each experiment and is depicted in \cref{tab:sampling_points} as an overview.

\todo{Higher number of sampling points? FOM has $3000^2 = 9 \text{Mio}$ sampling points.}
\begin{table}[htbp]
    \caption{Number of sampling points.}
    \begin{tabular}{|l|c|c|c|}
        \hline
        \textbf{experiment}
                          & $\mathbf{\ninterior}$
                          & $\mathbf{\nboundary}$
                          & $\mathbf{\nData}$                      \\
        \hline
        non-data-enriched & 30,000                & 3,000 & 0      \\
        \hline
        data-enriched     & 15,000                & 3,000 & 15,000 \\
        \hline
    \end{tabular}
    \centering
    \label{tab:sampling_points}
\end{table}

\subsection{Baseline: Non-Data-Enriched}
\label{sec:noDataPINN}
Before enriching the loss of the network with a data loss, we first establish a non-data-enriched PINN baseline, i.e.\ this experiment only features the interior loss $\linterior(\theta)$ and the boundary loss $\lboundary(\theta)$.\todo{Vorschlag, hinzufügen: We do not consider data of the solution.}


\begin{figure}[htbp]
    \centerline{\includegraphics[width=0.9\columnwidth]{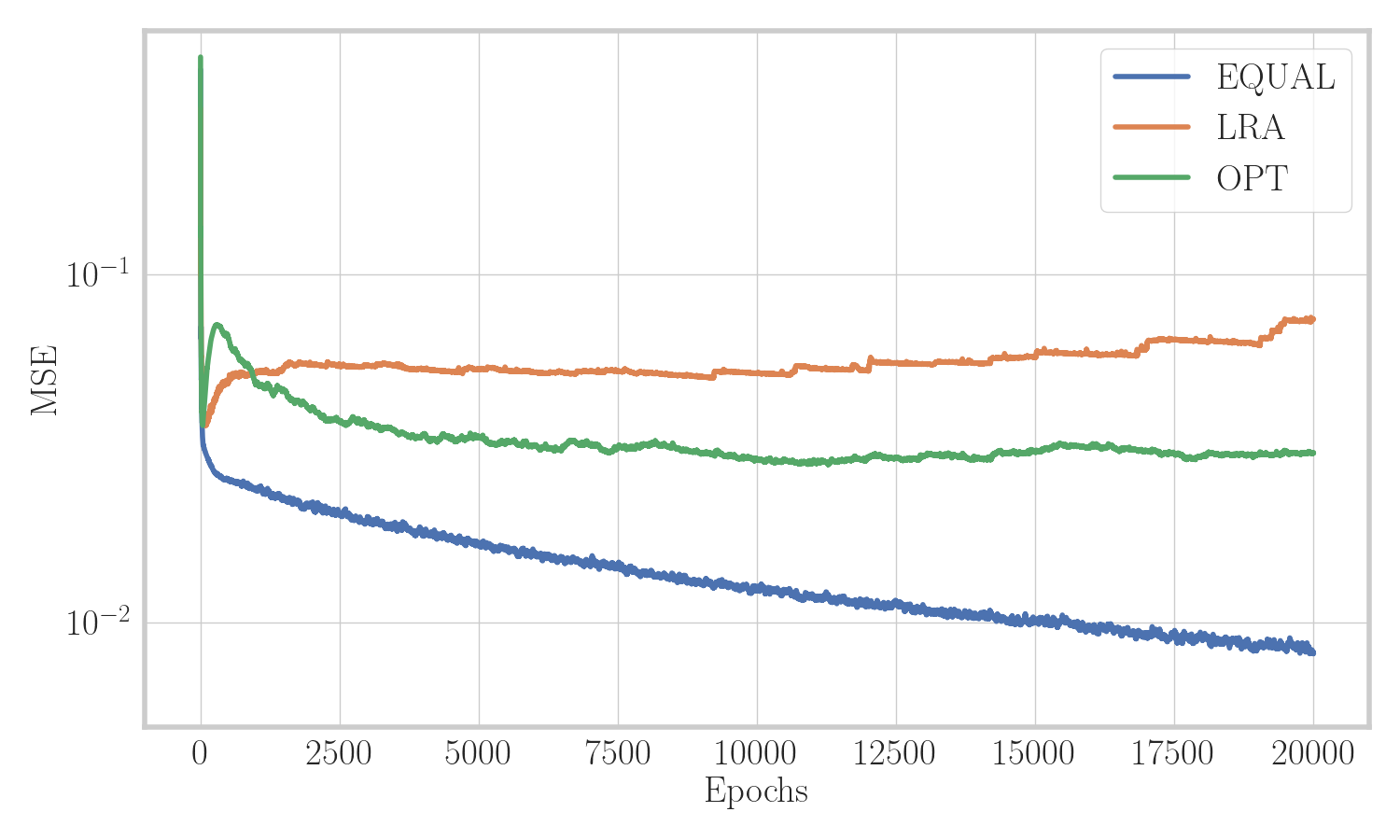}}
    \caption{Validation error $\mMSE{30}{\Phi}$ plotted over 20,000 training epochs of the non-data-enriched PINNs with EQUAL, LRA and OPT weighting.}
    \label{fig:noDataPINN}
\end{figure}

For the weighting of the loss contributions (see \cref{sec:weighting}), we consider equal weighting (EQUAL), i.e.\ $\lambda_j = 1$ for all $j \in \idxSet$, in addition to the LRA and OPT approach.

\cref{fig:noDataPINN} shows the training progress of all three weighting methods in terms of the validation error $\mMSE{30}{\fnet}$ over the number of epochs.
Neither of the three approaches was able to reliably capture the true solution, with minimum validation errors going only as low as $\num{1e-2}$.
This is in accordance with the observations in \cite{pinn_fail} that the one-dimensional wave equation is a very challenging problem for (non data-enriched) PINNs.
Note that the equal weighting approach performs best while the other, more informed, weightings result in bad outliers shifting the mean curve $\mMSE{30}{\fnet}$ upwards.
This certainly poses a strong case for including additional data during training as presented in the following.

\subsection{Data-Enriched: Exact Data}
\label{sec:data_enriched_explicit}
As a second baseline, we investigate how well data-enriched PINNs can train if the explicit solution is used in the data loss $\ldata$ from \cref{eq:ldata}. Note that this is a clearly unrealistic scenario but a good ``ideal'' method indicating the performance limits that we may expect with the ROM-data-enriched approaches.

\begin{figure}[htbp]
    \centerline{\includegraphics[width=0.9\columnwidth]{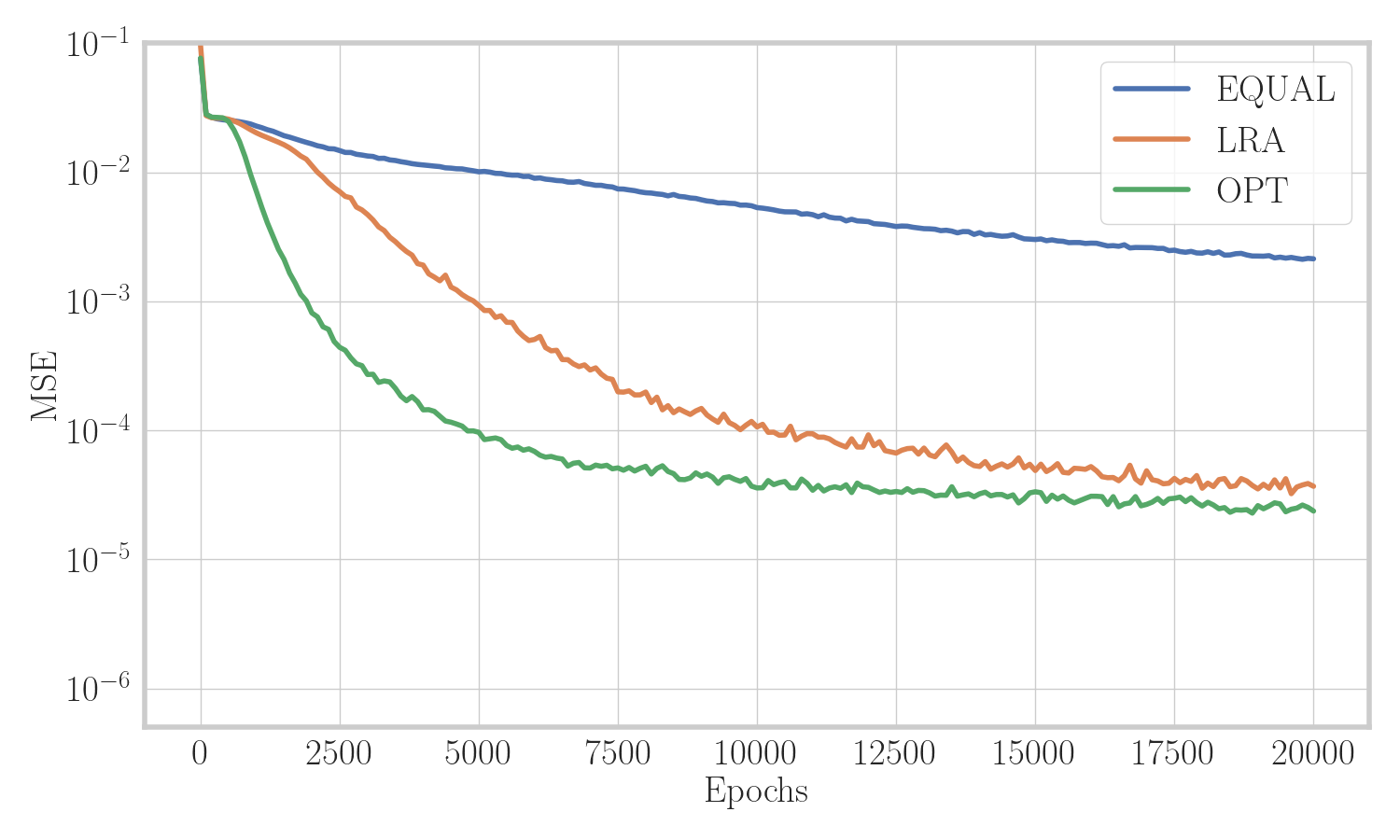}}
    \caption{Validation error $\mMSE{30}{\Phi}$ of the data-enriched PINNs (enriched with data of the explicit solution) plotted over 20,000 training epochs.}
    \label{fig:explicit_comparison}
\end{figure}

\cref{fig:explicit_comparison} shows the $\mMSE{30}{\Phi}$ for this data-enriched training for the three different weighting approaches.
The data-enrichment clearly improves the training performance as the validation error drops below $\num{1e-4}$. Additionally, it can clearly be seen that the OPT and LRA methods show a much quicker convergence behavior than EQUAL, with the OPT performing best overall.
Thus, we restrict the experiments in the following section to the OPT weighting.

\subsection{ROM-data-enriched PINNs}
Next, we replace the explicit solution in the data loss with a surrogate solution $\uROM$, i.e.\ $\yData = \uROM(\xData)$ in \cref{eq:loss_data}, where a ROM is used to compute the surrogate solution (see \cref{subsec:MOR}). To this end, the PDE is discretized with the Finite Element Method (Lagrangian elements on an equidistant grid, piecewise constant in time and piecewise linear in space, $3000$ discretization points in each dimension) which we refer to as Full-Order Model (FOM).
The error of the FOM is $\MSE{u_\text{FOM}}= \num{1.46e-6}$.
Based thereon, model order reduction is applied to derive three different ROMs of reduced sizes $n \in \{4, 8, 12 \}$ which varies the accuracy of the different ROMs.
The maximal dimension is set to $12$, which results in a reduction error of $\MSE{\uROM} = \num{3.85e-06}$.
Note that by construction of this experiment the ROMs are based on much more accurate data (FOM snapshots) than the data-enriched PINNs (only ROM snapshots).

\begin{figure}[htbp]
    \centerline{\includegraphics[width=0.9\columnwidth]{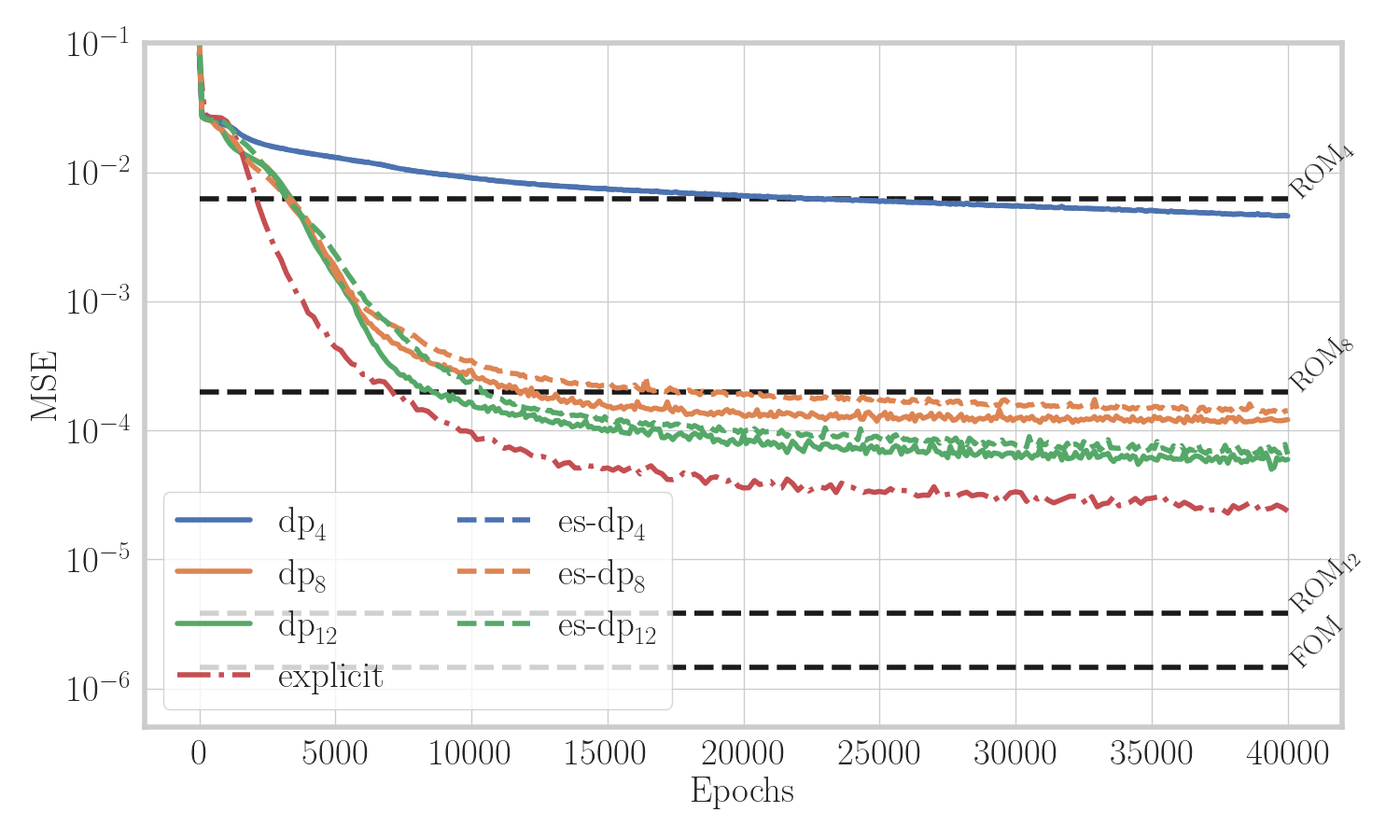}}
    \caption{Validation error $\mMSE{30}{\Phi}$
        of the (error-sensitive) ROM-data-enriched PINNs with the weighting methods OPT for three different ROM sizes $\{4, 8, 12\}$.
        Horizontal bars depict the respective error of the FOM and ROM solution.
    }
    \label{fig:overview}
\end{figure}

\todo{to drive home the argument, also plot non-data-enriched OPT in \cref{fig:overview}?}

\cref{fig:overview} shows the results for all six of these configurations, namely the ROM-data-enriched PINNs ($\textrm{dp}_n$, solid lines) and the error-sensitive variant ($\textrm{es-dp}_n$, dashed lines) for $n \in \{4, 8, 12 \}$.
For comparison purposes, the data-enriched PINN run using the explicit solution data from \cref{sec:data_enriched_explicit} is depicted (red line) and the MSE of the FOMs and ROMs are included as horizontal lines.
The graph shows that the ROM-data-enriched PINNs perform much better in terms of the overall predictive power when the ROM data is good enough ($n \ge 8$) achieving validation errors close to the model trained on explicit solution data.
Even more noteworthy, the ROM-data-enriched PINNs are able to improve over the error in the ROM for $n \le 8$. This can be seen as the respective MSE curves fall below the $\textrm{ROM}_4$ and $\textrm{ROM}_8$ markers.
The error-sensitive data-enrichment, however, does not improve the accuracy in this example.
This is assumed to be accounted to the fact that the non-data-enriched PINN model itself is not able to achieve a reasonable prediction as described in \cref{sec:noDataPINN}.




\section{Conclusion}
\label{seq:conclusion}
Our approach proved that it is beneficial to combine physics-aware neural networks with inexact data obtained from surrogate models.
We have shown that ROM-data-enriched PINNs can outperform both, conventional PINNs and ROMs.
The results presented here serve as a proof-of-concept, studying a problem for which the exact solution is known. We expect that error-sensitive PINNs will show their real benefit in higher-dimensional, parametrized simulation settings, which is subject of future work. In a parametric setting, it will be interesting to see how well the error-sensitive PINNs generalize to unseen parameters, avoiding prohibitively expensive simulation runs.
\todo{
    from a MOR point of few, a candidate for another model is the heat equation.
    Grepl estimator could be used (which is a space-time estimator, but increment time to get a time-dependet).
    Q: what is implemented in pyMOR?

    From PINN point of view, an existing model should be investigated for debugging.
}
\section*{Acknowledgment}
Funded by Deutsche Forschungsgemeinschaft (DFG, German Research Foundation) under Germany's Excellence Strategy - EXC 2075 – 390740016. We acknowledge the support by the Stuttgart Center for Simulation Science (SimTech).

\bibliographystyle{abbrv}
\bibliography{literature}

\end{document}

%% file: abbreviations.tex
\def\flayer{{f}}
\def\fnet{{\Phi}}

\newcommand{\nlayers}{n_l}
\newcommand{\nneurons}{n_n}

\newcommand{\diffOp}{\mathcal{N}}
\newcommand{\bdryOp}{\mathcal{B}}

\newcommand{\mabs}[1]{\left|#1\right|}
\newcommand{\mnorm}[1]{\left\lVert{#1}\right\rVert}
\newcommand{\LtNorm}[2]{\mnorm{#1}_{L_2(#2)}}
\newcommand{\LtDomainNorm}[1]{\LtNorm{#1}{\domain}}

\newcommand{\R}{\mathbb{R}}

\newcommand{\lb}{\left(}
\newcommand{\rb}{\right)}

\newcommand{\intd}{\text{d}}

\def\XXint#1#2#3{{\setbox0=\hbox{$#1{#2#3}{\int}$}
     \vcenter{\hbox{$#2#3$}}\kern-.5\wd0}}

\newcommand{\domain}{\varOmega}
\newcommand{\domainXi}{\domain_\xi}

\newcommand{\domainT}{\domain_t}
\newcommand{\dxi}{\Delta_\xi}
\newcommand{\dt}{\Delta_t}

\newcommand{\nT}{n_t}
\newcommand{\nXi}{n_\xi}
\newcommand{\LtDomainXiNorm}[1]{\LtNorm{#1}{\domainXi}}

\newcommand{\uROM}{\widetilde{u}}

\newcommand{\nTheta}{n_\theta}
\newcommand{\loss}{\mathcal{L}}
\newcommand{\approxLoss}{l}
\newcommand{\dataLabel}{\text{D}}
\newcommand{\Ldata}{\loss_\dataLabel}
\newcommand{\Mdata}{M_\dataLabel}
\newcommand{\ldata}{\approxLoss_\dataLabel}
\newcommand{\nData}{n_\dataLabel}

\newcommand{\xData}[1][i]{x_{\dataLabel, #1}}
\newcommand{\yData}[1][i]{y_{\dataLabel, #1}}
\newcommand{\lambdaData}{\lambda_\dataLabel}
\newcommand{\interiorLabel}{\text{I}}

\newcommand{\linterior}{\approxLoss_\interiorLabel}

\newcommand{\ninterior}{n_\interiorLabel}
\newcommand{\xInterior}[1][i]{x_{\interiorLabel, #1}}
\newcommand{\boundaryLabel}{\text{B}}

\newcommand{\lboundary}{\approxLoss_\boundaryLabel}
\newcommand{\xBoundary}[1][i]{x_{\boundaryLabel, #1}}

\newcommand{\nboundary}{n_\boundaryLabel}
\newcommand{\idxSet}{\mathcal{J}}

\newcommand{\errBound}{\varepsilon}
\newcommand{\ixi}{i_\xi}
\newcommand{\ball}[1][t]{B_\varepsilon(\uROM; #1)}
\newcommand{\closedBall}{\overline{B}_\varepsilon(\uROM; t)}
\newcommand{\ReLu}{\texttt{ReLu}}

\newcommand{\LdataWrom}{\loss_{\dataLabel, \text{ES}}}
\newcommand{\IxidataWrom}{I_{\xi, \text{ES}}}
\newcommand{\ixidataWrom}{i_{\xi, \text{ES}}}

\newcommand{\LdataHeur}{\loss_{\dataLabel, \text{ES}}}
\newcommand{\ldataHeur}{\approxLoss_{\dataLabel, \text{ES}}}

\newcommand{\MSE}[1]{\text{MSE}[#1]}
\newcommand{\mMSE}[2]{\overline{\text{MSE}}_{#1}[#2]}

%% file: ms.bbl
\begin{thebibliography}{10}

\bibitem{Baydin2018}
A.~G. Baydin, B.~A. Pearlmutter, A.~A. Radul, and J.~M. Siskind.
\newblock Automatic differentiation in machine learning: a survey.
\newblock {\em Journal of Machine Learning Research}, 18(153):1--43, 2018.

\bibitem{Glas2020}
S.~Glas, A.~T. Patera, and K.~Urban.
\newblock A reduced basis method for the wave equation.
\newblock {\em International Journal of Computational Fluid Dynamics},
  34(2):139--146, 2020.

\bibitem{init}
X.~Glorot and Y.~Bengio.
\newblock Understanding the difficulty of training deep feedforward neural
  networks.
\newblock In {\em Proceedings of the Thirteenth International Conference on
  Artificial Intelligence and Statistics}, 2010.

\bibitem{nn_pde_data}
Y.~Khoo, J.~Lu, and L.~Ying.
\newblock Solving parametric {PDE} problems with artificial neural networks.
\newblock {\em European Journal of Applied Mathematics}, page 1–15, 2020.

\bibitem{adam}
D.~P. Kingma and J.~Ba.
\newblock Adam: {A} method for stochastic optimization.
\newblock In Y.~Bengio and Y.~LeCun, editors, {\em 3rd International Conference
  on Learning Representations, {ICLR} 2015, San Diego, CA, USA, May 7-9, 2015,
  Conference Track Proceedings}, 2015.

\bibitem{Koppel2019}
M.~Köppel, F.~Franzelin, I.~Kröker, S.~Oladyshkin, G.~Santin, D.~Wittwar,
  A.~Barth, B.~Haasdonk, W.~Nowak, D.~Pflüger, and C.~Rohde.
\newblock Comparison of data-driven uncertainty quantification methods for a
  carbon dioxide storage benchmark scenario.
\newblock {\em Computational Geosciences}, 23(2):339--354, Apr. 2019.

\bibitem{nn_pde1}
I.~E. {Lagaris}, A.~{Likas}, and D.~I. {Fotiadis}.
\newblock Artificial neural networks for solving ordinary and partial
  differential equations.
\newblock {\em IEEE Transactions on Neural Networks}, 9(5):987--1000, 1998.

\bibitem{neural_op}
Z.~Li, N.~B. Kovachki, K.~Azizzadenesheli, B.~Liu, K.~Bhattacharya, A.~M.
  Stuart, and A.~Anandkumar.
\newblock Fourier neural operator for parametric partial differential
  equations.
\newblock In {\em 9th International Conference on Learning Representations,
  {ICLR} 2021, Virtual Event, Austria, May 3-7, 2021}. OpenReview.net, 2021.

\bibitem{dl_lstm}
A.~T. Mohan and D.~V. Gaitonde.
\newblock A deep learning based approach to reduced order modeling for
  turbulent flow control using {LSTM} neural networks.
\newblock {\em arXiv:1804.09269}, 2018.

\bibitem{nn_pde_cnn}
O.~Obiols-Sales, A.~Vishnu, N.~Malaya, and A.~Chandramowliswharan.
\newblock {CFDNet}: A deep learning-based accelerator for fluid simulations.
\newblock In {\em Proceedings of the 34th ACM}, ICS '20, New York, NY, USA,
  2020. Association for Computing Machinery.

\bibitem{nn_pde2}
D.~C. Psichogios and L.~H. Ungar.
\newblock A hybrid neural network-first principles approach to process
  modeling.
\newblock {\em AIChE Journal}, 38(10):1499--1511, 1992.

\bibitem{pinn}
M.~Raissi, P.~Perdikaris, and G.~Karniadakis.
\newblock Physics-informed neural networks: A deep learning framework for
  solving forward and inverse problems involving nonlinear partial differential
  equations.
\newblock {\em Journal of Computational Physics}, 378:686--707, 2019.

\bibitem{pinn_hfm}
M.~Raissi, A.~Yazdani, and G.~E. Karniadakis.
\newblock Hidden fluid mechanics: Learning velocity and pressure fields from
  flow visualizations.
\newblock {\em Science}, 367(6481):1026--1030, 2020.

\bibitem{pinn_subsurface}
A.~M. Tartakovsky, C.~O. Marrero, P.~Perdikaris, G.~D. Tartakovsky, and
  D.~Barajas-Solano.
\newblock Physics-informed deep neural networks for learning parameters and
  constitutive relationships in subsurface flow problems.
\newblock {\em Water Resources Research}, 56(5), 2020.

\bibitem{biomech}
J.~Valentin, M.~Sprenger, D.~Pflüger, and O.~Röhrle.
\newblock Gradient-based optimization with {B}-splines on sparse grids for
  solving forward-dynamics simulations of three-dimensional,
  continuum-mechanical musculoskeletal system models.
\newblock {\em International Journal for Numerical Methods in Biomedical
  Engineering}, 34(5):e2965, 2018.

\bibitem{vdM2021}
R.~van~der Meer, C.~Oosterlee, and A.~Borovykh.
\newblock Optimally weighted loss functions for solving {PDE}s with neural
  networks.
\newblock {\em arXiv:2002.06269}, 2021.

\bibitem{lra2020}
S.~Wang, Y.~Teng, and P.~Perdikaris.
\newblock Understanding and mitigating gradient pathologies in physics-informed
  neural networks.
\newblock {\em arXiv:2001.04536}, 2020.

\bibitem{pinn_fail}
S.~Wang, X.~Yu, and P.~Perdikaris.
\newblock When and why pinns fail to train: A neural tangent kernel
  perspective.
\newblock {\em Journal of Computational Physics}, page 110768, 2021.

\end{thebibliography}
